\documentclass[a4paper,conference]{IEEEtran}

\IEEEoverridecommandlockouts
\usepackage{cite}
\usepackage{amsmath,amssymb,amsfonts}
\usepackage{graphicx}
\usepackage{textcomp}
\usepackage{xcolor}
\usepackage{amsmath}
\usepackage{multirow}
\usepackage{multicol}
\usepackage{caption}
\usepackage{array}
\usepackage{subcaption}
\usepackage{balance}
\usepackage{wrapfig}
\usepackage{hyperref}
\usepackage{multirow}
\usepackage{multicol}
\usepackage{color}
\usepackage{parskip}
\usepackage{colortbl}
\usepackage{booktabs}
\usepackage{flushend}
\usepackage{hyperref}
\usepackage{todonotes}
\usepackage{comment}
\usepackage{adjustbox}
\usepackage{mathtools}
\usepackage{colortbl}
\usepackage{placeins}
\usepackage{adjustbox}
\usepackage{lipsum,booktabs}
\usepackage{parskip}
\usepackage{tabularx,ragged2e,booktabs}
\usepackage{subcaption}
\newcolumntype{L}{>{\RaggedRight\arraybackslash}X}

\usepackage{caption}
\usepackage{environ} 
\NewEnviron{NORMAL}{%
    \scalebox{0.80}{$\BODY$} 
} 
 
\NewEnviron{HUGE}{%
    \scalebox{5}{$\BODY$} 
} 
\newcolumntype{L}{>{\centering\arraybackslash}m{3cm}}

\usepackage{algpseudocode}
\usepackage[ruled]{algorithm2e} 
\newcolumntype{L}{>{\centering\arraybackslash}m{3cm}}

\def\BibTeX{{\rm B\kern-.05em{\sc i\kern-.025em b}\kern-.08em
    T\kern-.1667em\lower.7ex\hbox{E}\kern-.125emX}}
    
\makeatletter
\newcommand{\linebreakand}{%
  \end{@IEEEauthorhalign}
  \hfill\mbox{}\par
  \mbox{}\hfill\begin{@IEEEauthorhalign}
}
\makeatother

\begin{document}

\title{Entity-driven Fact-aware Abstractive Summarization of Biomedical Literature\\

}

\author{
    \IEEEauthorblockN{Amanuel Alambo\IEEEauthorrefmark{1},
    Tanvi Banerjee\IEEEauthorrefmark{1},
    Krishnaprasad Thirunarayan\IEEEauthorrefmark{1},
    Michael Raymer\IEEEauthorrefmark{1}}
    \IEEEauthorblockA{\IEEEauthorrefmark{1}Wright State University
    \\\{alambo.2, tanvi.banerjee, t.k.prasad, michael.raymer\}@wright.edu}
}

\newcommand{\tb}[1]{\textcolor{purple}{{\tiny{TB: }}#1}}


\maketitle

\begin{abstract}
As part of the large number of scientific articles being published every year, the publication rate of biomedical literature has been increasing. Consequently, there has been considerable effort to harness and summarize the massive amount of biomedical research articles. While transformer-based encoder-decoder models in a vanilla source document-to-summary setting have been extensively studied for abstractive summarization in different domains, their major limitations continue to be entity hallucination (a phenomenon where generated summaries constitute entities not related to or present in source article(s)) and factual inconsistency. This problem is exacerbated in a biomedical setting where named entities and their semantics (which can be captured through a knowledge base) constitute the essence of an article. The use of named entities and facts mined from background knowledge bases pertaining to the named entities to guide abstractive summarization has not been studied in biomedical article summarization literature. In this paper, we propose an entity-driven fact-aware framework for training end-to-end transformer-based encoder-decoder models for abstractive summarization of biomedical articles. We call the proposed approach, whose building block is a transformer-based model, EFAS, Entity-driven Fact-aware Abstractive Summarization. We conduct a set of experiments using five state-of-the-art transformer-based encoder-decoder models (two of which are specifically designed for long document summarization) and demonstrate that injecting knowledge into the training/inference phase of these models enables the models to achieve significantly better performance than the standard source document-to-summary setting in terms of entity-level factual accuracy, N-gram novelty, and semantic equivalence while performing comparably on ROUGE metrics. The proposed approach is evaluated on ICD-11-Summ-1000, a dataset we build for abstractive summarization of biomedical literature, and PubMed-50k, a segment of a large-scale benchmark dataset for abstractive summarization of biomedical literature.
\end{abstract}


\section{Introduction}
Neural abstractive summarization is well explored for summarization of news articles \cite{rush2015neural, chopra2016abstractive, see2017get, fabbri2019multi, liu2019text}, and scientific articles \cite{cohan2018discourse, sharma2019bigpatent, cachola2020tldr}. While there are some efforts to apply neural abstractive summarization techniques for biomedical literature \cite{cohan2018discourse, esteva2020co}, the use of named entities and background knowledge bases to guide biomedical abstractive summary generation has not been explored. On the other hand, named entity recognition/understanding in biomedical literature has been extensively studied such that named entities are known to harbor significant semantics about a biomedical article \cite{dougan2014ncbi, li2016biocreative, zhao2017disease, kocaman2021biomedical}. Further, linking named entities to concept definitions in background domain-specific knowledge bases boost semantic understanding of a biomedical article by improving comprehensiveness and contextualization as investigated in \cite{shen2014entity, zheng2015entity, zhu2020latte}.

The use of named entities to guide abstractive summarization of news articles has been explored in recent studies by \cite{sharma2019entity, zhou2021entity, narayan2021planning, gunel2020mind}. However, these studies do not leverage named entity-driven facts from background knowledge bases to guide abstractive summary generation. Inspired by recent advances in transformer-based encoder-decoder models \cite{vaswani2017attention, lewis2019bart, raffel2020exploring, zhang2020pegasus} and knowledge and retrieval augmented natural language generation \cite{zhang2019ernie, lewis2020retrieval, guu2020realm, karpukhin2020dense, petroni2020kilt, an2021retrievalsum}, we propose a technique for retrieval of entity-driven facts from biomedical knowledge bases and leveraging the retrieved facts as contextual signals for abstractive summarization of biomedical literature. Our proposed framework consists of two major components: 1) entity-driven knowledge retriever; and 2) knowledge-guided abstractive summarizer. The entity-driven knowledge retriever extracts facts from UMLS \cite{bodenreider2004unified}, ICD-10 \cite{hirsch2016icd}, and SNOMED-CT \cite{donnelly2006snomed} based on named entities in a biomedical article while the knowledge-guided abstractive summarizer is trained to generate summaries by attending to the source article to be summarized, the chain of named entities in the source article, and the retrieved facts from the knowledge bases. 

We conduct experiments on two datasets: ICD-11-Summ-1000, and PubMed-50k. For curating ICD-11-Summ-1000, we conduct entity-driven clustering of PubMed abstracts collected per ICD-11 \cite{harrison2021icd} chapter followed by entity-aware content selection to build an entity-aware pseudo extractive document. The pseudo extractive document is passed as an input source article to generate the abstractive summary during inference. The contributions of this study are summarized as follows:
\begin{itemize}

\item We introduce an approach for named entity-driven fact retrieval from biomedical knowledge bases using dense vector representations.

\item We propose a framework based on transformer-based encoder-decoder models for knowledge-guided large-scale abstractive summarization of biomedical literature. 

\item We conduct extensive experiments and ablation studies to assess the efficacy of the proposed approach and show that injecting named entities and entity-aware facts mined from biomedical knowledge bases into abstractive summarization models can boost their performance in terms of entity-level factual consistency \cite{goodrich2019assessing, kryscinski2019evaluating, narayan2021planning, nan2021entity}, N-gram novelty \cite{an2021retrievalsum, zhong2020extractive}, and semantic equivalence measured using BERTScore \cite{zhang2019bertscore}.

\item We develop ICD-11-Summ-1000, a dataset consisting of clusters of biomedical articles collected from PubMed for the \emph{special groups} chapters in the ICD-11 catalog \footnote{https://bit.ly/3GEFWvM} and the abstractive summaries generated using different variations of our experimental design. Further, we plan to share our code, and data with other researchers \footnote{\url{https://github.com/AmanuelF/biomed_abstractive_summarization}}.
\end{itemize}

\section{Related Work}
While abstractive summarization is well studied for summarization of news articles with success attributed to the availability of a massive amount of training data, their applicability to scholarly articles, particularly, in the biomedical domain is limited. Further, although named entities have been extensively studied to convey the semantics of an article (news, scientific, social media) and the saliency of individual sentences \cite{zhou2021entity} within an article, they have not been widely used as part of modeling abstractive summarization. \cite{sharma2019entity} performed entity-aware single-document abstractive summarization using reinforcement learning for training. Their pipeline-based approach consists of an entity-aware content selection module and abstract generation module. They evaluate their approach on the CNN/Daily Mail and NYT corpora. \cite{zhou2021entity} perform entity-driven multi-document abstractive summarization of news articles (WikiSum, and Multi-News) using an encoder-decoder framework augmented with Graph Attention Network (GAT). \cite{schulze2016entity} proposed EntityRank, an extension of the LexRank \cite{erkan2004lexrank} graph-based algorithm, for entity-supported summarization of biomedical abstracts. 

There have been a few recent efforts towards knowledge/fact-aware abstractive summarization in different domains. \cite{zhu2020enhancing} introduced a Fact-aware Abstractive Summarization model called FaSum for improving the factual consistency of summaries in the domain of news articles. However, their approach does not leverage named entities for fact retrieval. \cite{gunel2020mind} extended a transformer-based abstractive summarization model using entities disambiguated and linked to Wikidata knowledge graph and attending to the entities for summarization of news articles. Their approach, however, does not perform named-entity based fact retrieval from the knowledge base constrained by the article to be summarized and the named entities. \cite{manas2021knowledge} developed an unsupervised pipeline-based approach for knowledge-infused abstractive summarization for condensing patient-to-clinician diagnostic interviews based on Multi-Sentence Compression \cite{filippova2010multi} and Integer Linear Programming \cite{schrijver1998theory}. Nevertheless, their approach uses domain-specific lexicons as knowledge source for filtering irrelevant utterances and for retroffiting language models \cite{faruqui2014retrofitting} and, does not leverage named entities or facts as part of an end-to-end training of models. \cite{afzal2020clinical} proposed Biomed-Summarizer, a framework for extractive summarization of biomedical literature in a multi-document setting and evaluated on PubMed abstracts.  \cite{cohan2018discourse} built a model for abstractive summarization of long documents using a discourse-aware encoder-decoder framework and experimented on two large scale datasets including research articles collected from PubMed. To address the challenge associated with the scarcity of large-scale training data in the biomedical domain, \cite{deyoung2021ms2} released MS2 (Multi-Document Summarization of Medical Studies). They experimented with BART \cite{lewis2019bart} for abstractive summary generation on the dataset they introduced in a traditional multi-doc-to-summary setting. 

Though all the aforementioned studies conduct abstractive summarization of biomedical literature or the use of facts mined from knowledge bases for a different domain, they follow the well-established paradigm of source-document vs summary pairing  during training/inference of models. Our approach is different in that we augment the state-of-the-art abstractive summarization models with additional contextual signals during training/inference and apply them to the biomedical domain.

\section{Data Preparation}
In addition to the benchmark PubMed-50k \cite{cohan2018discourse} dataset which we use to train our models, we curate ICD-11-Summ-1000 dataset using a data preparation pipeline which follows two stages: 1) ICD-11 disease lexicon curation for querying PubMed for abstracts; and 2) Entity-aware pseudo-document generation for a collection of semantically related PubMed abstracts in an ICD-11 chapter. Thus, for querying PubMed for abstracts for an ICD-11 chapter, we first query the biomedical knowledge bases for “disease” keywords using the names of each ICD-11 chapter and build a lexicon of diseases corresponding to each chapter. Figure-1 shows what a lexicon build-up for an ICD-11 chapter looks like. Once the disease related keywords are identified for an ICD-11 chapter, we use these keywords to query PubMed via the Bio Entrez parser \footnote{https://biopython.org/docs/1.75/api/Bio.Entrez.html} to capture the first 1000 abstracts (PMIDs) spanning a period of last 90 days from the moment we initiated the query. We do this for each of the eight \emph{special groups} ICD-11 chapters.

\begin{figure}[h]
  \centering
  \captionsetup{font=scriptsize}
  \includegraphics[width=0.45\textwidth]{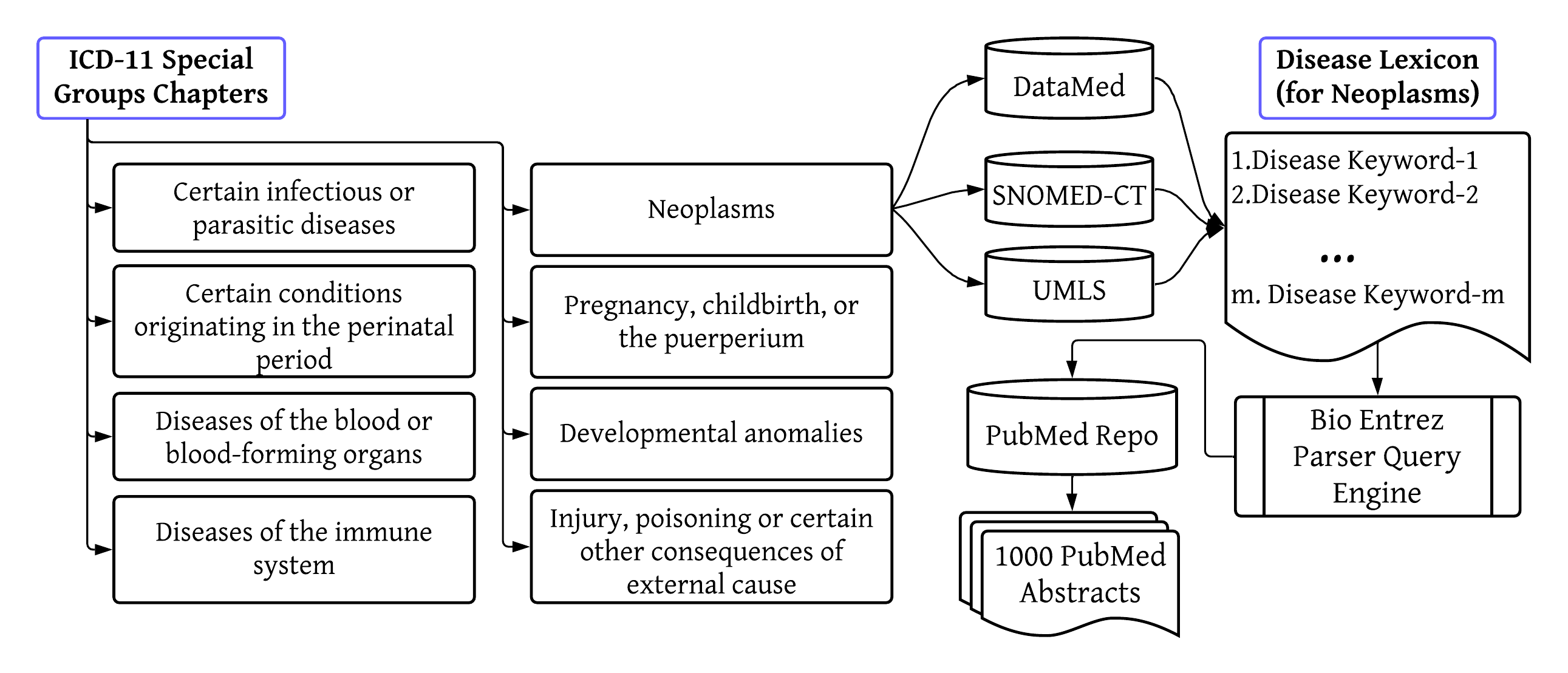}
  \caption{ ICD-11 based lexicon construction and querying for abstracts from PubMed using Bio Entrez parser.  For illustration purpose, we show the pipeline for ICD-11 chapter \emph{Neoplasms}}
\vspace{-4mm}
\end{figure}

Figure-2 shows our ICD-11-Summ-1000 dataset preparation pipeline. Once we have queried the 1000 PubMed abstracts for an ICD-11 chapter, we conduct named entity recognition (NER) on each of the abstracts within a chapter using the SciSpacy NER model trained using the BC5CDR corpus \cite{neumann2019scispacy}. Since we are interested in entity-level clustering of PubMed abstracts within an ICD-11 chapter, we first conduct clustering of the named entities using agglomerative clustering as used in \cite{zhou2021entity}. We use BioBERT \cite{lee2020biobert} for named entity representation followed by agglomerative clustering. Once the named entities pertaining to an ICD-11 chapter are clustered into different bins, our next task is to cluster the PubMed abstracts into a bin based on how the named entities within the abstracts are related to the entities within a cluster. We use cosine similarity between named entities identified in a PubMed abstract and entities characterizing a cluster to determine an entity-aware cluster the abstract belongs to. Next, for each cluster, we perform named entity-aware salient content selection to produce an extractive pseudo-document for each cluster. This paradigm of reducing a multi-document corpus (i.e., a cluster consisting of PubMed abstracts grouped based on entity-relatedness) into an extractive pseudo-document is explored for different tasks in previous studies \cite{lebanoff2018adapting, zhang2018towards, gao2020supert}. As part of the NER task, we use coreference resolution \cite{lee2017end, gardner2018allennlp} after named entities are extracted using SciSpacy to cluster the biomedical named entities and their coreferenced mentions spanning the multiple abstracts within an ICD-11 chapter.

\begin{figure}[h]
  \centering
  \captionsetup{font=scriptsize}
  \includegraphics[width=0.43\textwidth]{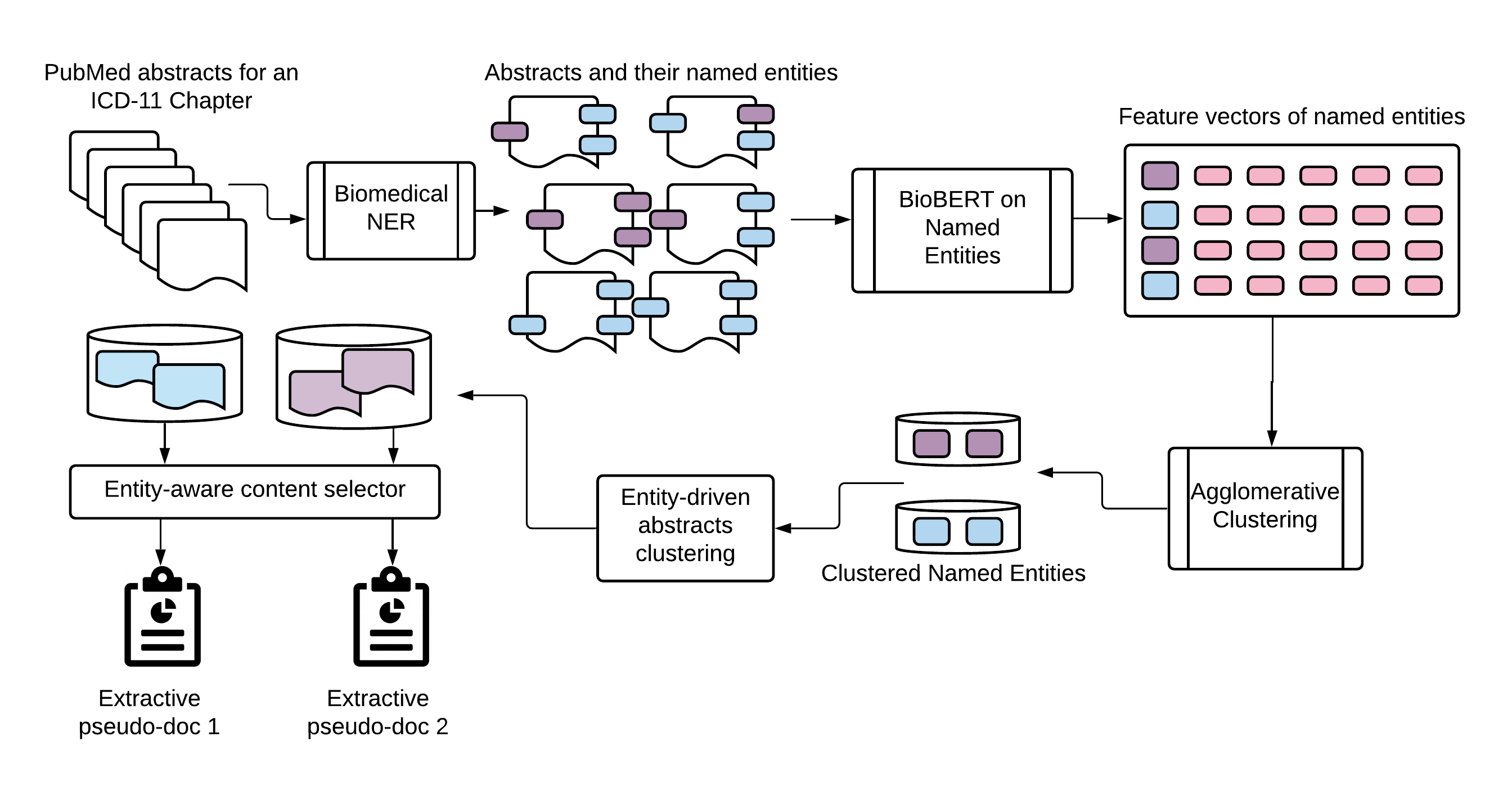}
  \caption{Entity-aware content selection to produce extractive pseudo-documents. The light blue and light lavender colored documents in the final bins represent abstracts whose named entities are semantically similar to one another.}
\vspace{-4mm}
\end{figure}

During entity-aware content selection to produce an extractive pseudo-doc for a cluster of PubMed abstracts that are clustered based on named entity relatedness, we preserve the positioning of sentences within an abstract. We also use the following heuristics while constructing the extractive pseudo-doc: 1) a sentence shall have at least one named entity identified using SciSpacy-BC5CDR NER model; and 2) the selected sentences from an abstract are placed in the same order as they appear in the abstract. Further, we also take into account abstracts’ relative importance scores where abstracts with higher document importance scores \cite{banerjee2015multi} have their sentences precede sentences from abstracts with less document importance scores while generating the extractive pseudo-document. Document importance $D_{imp}$ of target abstract $d_i$ is determined using pairwise cosine similarity between the BioBERT \cite{lee2020biobert} embedding of $d_i$ and other abstracts within the same cluster $\mathcal C$. Formally, document importance is defined as

\begin{equation}
    D_{imp} = \frac{\sum_{d_i, d_j \in \mathcal C} cossim(d_i, d_j) }{|\mathcal C|-1}, (i \neq j)
\end{equation}

For all tasks throughout this paper involving initializing of networks or for representation learning, we use BioBERT \cite{lee2020biobert}. 

\section{Proposed Method}
Our proposed approach is a two-stage framework consisting of 1) an entity-driven knowledge retriever, and 2) a knowledge-guided abstractive summarizer. In this section, we discuss both modules in detail. 

\subsection{Entity-driven Knowledge Retriever}
For each extractive pseudo-document generated in the data preparation stage for ICD-11 or input article for PubMed-50k designated by $\mathcal{D}$, we identify the named entities in the input document. The identified named entities are then used to retrieve facts from biomedical knowledge bases (UMLS, ICD-10, and SNOMED-CT). We use PyMedTermino \cite{jean2021using} to work with the entire dump of UMLS \cite{bodenreider2004unified} available at \footnote{https://bit.ly/3E0zrll}. For $m$ named entities (and their coreferenced mentions), we have a set of pairs of entities  \{${(e_i , e_j) ~ | ~ 0 \leq i < j < m }$\} 
extracted from $\mathcal{D}$, where each pair $(e_i , e_j)$ is used to query for $c$ candidate facts ${{F}_1, {F}_2, {F}_3 , \ldots , {F}_{|c|}}$ denoted collectively by ${F}^{i, j}_{\mathcal{D}}$ from the background knowledge bases $\mathcal{K}$  using full text search. The complete set of facts retrieved for all pairs of named entities in source document $\mathcal{D}$ is denoted by ${\mathcal{F}}_{\mathcal{D}}$.

The reason we use a pair of named entities to perform lexical query from $\mathcal{K}$ is to capture the relationship between a pair of named entities as it appears in a knowledge base since relationships among named entities harbor semantics in addition to the entities themselves and help with disambiguating relevant facts from irrelevant facts. After the candidate facts ${\mathcal{F}}_{\mathcal{D}}$ are retrieved from the knowledge bases $\mathcal{K}$, we embed the candidate facts using BioBERT. Then, we perform efficient vector similarity search using Maximum Inner Product Search (MIPS) \cite{shrivastava2014asymmetric} implemented in the FAISS library \footnote{https://github.com/facebookresearch/faiss} to query for the top-k facts among the candidate facts (${\mathcal{F}}_\mathcal{D}$) using the input document $\mathcal{D}$ as the query. Formally, we define the similarity between fact ${F}_i \in {\mathcal{F}}_\mathcal{D}$ and document $\mathcal{D}$ as

\vspace{-3mm}

\begin{equation}
    sim( F_i, \mathcal D) = \vec{\mathcal V} ( F_i)^T \vec{\mathcal V} (\mathcal D)
\end{equation}

\vspace{-2mm}

where $\vec{\mathcal V} (F_i) $ - Vector representation of Fact $F_i$; $\vec{\mathcal V} (\mathcal D) $ - Vector representation of document $\mathcal D$ 

Thus, after the knowledge retrieval task, we have 1) the input document $\mathcal{D}$ which is obtained during the data preparation phase for ICD-11 and readily available for PubMed-50k; 2) the named entity chain (i.e., chain of named entities extracted from the pseudo-doc) $\mathcal{E}$ \cite{narayan2021planning}; and 3) top-k facts ${{F}_1, {F}_2, {F}_3, \ldots , {F}_{|K|}}$ retrieved from the background knowledge bases collectively represented as ${F}_K \subseteq {\mathcal{F}}_\mathcal{D}$. We set the value of $K$ to 3 following the study by \cite{an2021retrievalsum}. We experiment with different values of $K$ as detailed in the ablation studies section. The combination of these contextual signals will be used to guide the summarization model at training/inference time. The rationale for using maximum inner product search for knowledge retrieval is inspired by the works of \cite{karpukhin2020dense, lewis2020retrieval, guu2020realm, izacard2020leveraging, singh2021end}, albeit they used it mainly for open domain question answering \cite{prager2006open, lewis2020question}.  \cite{an2021retrievalsum} use a similar approach for exemplar retrieval in their RetrievalSum model which is based on contrastive learning \cite{chopra2005learning} using a Siamese network \cite{koch2015siamese} to learn representations for an input document and the exemplars and guide their summary generation. Our problem of retrieving the most relevant facts from the background KB, however, is framed as a dense passage retrieval problem. Named entities from the input document are extracted using the SciSpacy NER model trained on the BC5CDR corpus \cite{neumann2019scispacy}. Table-I shows sample facts, as they appear and retrieved from UMLS KB for an input article with a given pair of named entities identified.

\begin{table}[htbp]
\centering
\captionsetup{font=scriptsize}
\begin{center}
\scalebox{0.70}{
\begin{tabular}
{| p{3.0cm} | p{7.75cm}|}
\hline

\cline{2-2} 

\textbf{\textbf{Named Entity Pair}} & \textbf{\textbf{Entity-driven Facts from UMLS KB}}  \\

\hline
\emph{(iron, anemia)} & \emph{Iron} deficiency \emph{anemia} secondary to inadequate dietary \emph{iron} intake. \newline
\emph{Iron} deficiency \emph{anemia} in mother complicating childbirth.\\

\hline
\emph{(dementia, depression)} & Primary degenerative \emph{dementia} of the Alzheimer type, presenile onset, with \emph{depression}. \newline
Arteriosclerotic \emph{dementia} with \emph{depression}. \\

\hline
\emph{(diabetes, hypertension)} & \emph{Hypertension} in chronic kidney disease due to type 1 \emph{diabetes} mellitus. \newline 
\emph{Hypertension} concurrent and due to end stage renal disease on dialysis due to type 2 \emph{diabetes} mellitus. \\
\hline

\end{tabular}
}
\label{tab1}
\end{center}
\vspace{-1mm}
\caption{Pairs of named entities and sample facts mined from UMLS for each pair.}
\end{table}

\vspace{-2mm}

\subsection{Knowledge-guided Abstractive Summarizer}
The backbone component of our knowledge-guided abstractive summarizer, which is a transformer encoder-decoder model, is based on the work by \cite{dou2020gsum}. Figure-3 shows the proposed end-to-end model architecture. We use this architectural setup for all the models we experiment with. We designate a model augmented with one of the knowledge signals as model-EFAS. We train the models on the 50k samples obtained from PubMed abstractive scientific summarization dataset \cite{cohan2018discourse} using different combinations of signals (with and without named entities and facts). The top-k facts retrieved by the biomedical knowledge retriever, corresponding to each pair of named entities in an input extractive pseudo-doc or input article, are separated by a special token \textbf{[SEP]}. The input article is passed as one input document prepended with \textbf{[CLS]} and appended with \textbf{[SEP]} token.  The named entity chain is passed as one segment prepended with \textbf{[CLS]} and appended with \textbf{[SEP]} token. There have been different approaches to combining different signals such as concatenating the different pieces to prime the generation component such as the one proposed in Fusion-in Decoder \cite{izacard2020leveraging} and \cite{singh2021end}. The top-k retrieved facts are initialized using BioBERT and the concatenated encoding is then passed through a sequence of transformer layers to be projected onto a 768- dimension vector to later be attended to by the autoregressive decoder. Similarly, the named entity chain is initialized with BioBERT and passed through a sequence of transformer encoders. Each transformer encoder layer is composed of self-attention and feed forward sub-layers. At training time, a batch of input-output pairings is passed to the encoder and decoder respectively in the form $\langle x, y \rangle$. The encoder undergoes the following transformations to the input sequence $x$ which in this formalism is used to represent the first hidden layer $h^0$ of the stacked sequence of $l$ transformer encoder layers.

\begin{center}
$\tilde{h}^l_x := LayerNorm(h^{l-1}_x + MHAtt(h^{l-1}_x))$ \newline
$h^l_x := LayerNorm(\tilde{h}^l_x + FFN(\tilde{h}^l_x))$ \newline
\end{center}

\vspace{-2mm}

\begin{figure}[h]
  \centering
  \captionsetup{font=scriptsize}
  \includegraphics[width=0.45\textwidth]{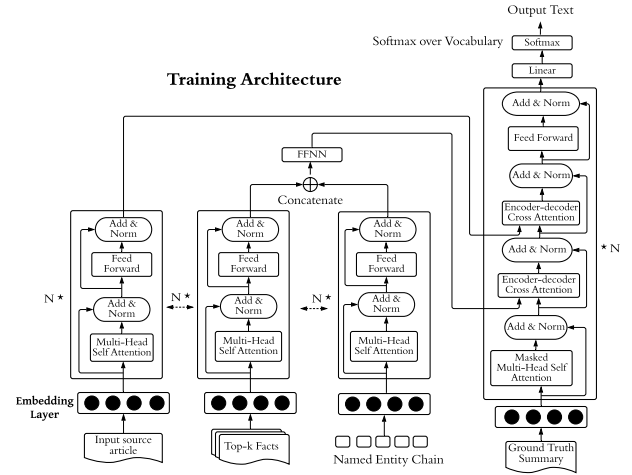}
  \caption{The Proposed Framework. The encoder networks have their parameters shared. The two cross attention sub-layers in the decoder attend to the input source article, and a linear transformed projection of encodings of facts, and the chain of named entities. This architecture best represents the three traditional transformer models. For BigBird, and LED, the full self attention layer gets replaced with sparse attention.}
\vspace{-4mm}
\end{figure}

The decoder component, which is trained using teacher forcing \cite{williams1989learning} at training time, consists of two cross-attention sub-layers to attend to: 1) the input source article; and 2) the affine transformed concatenation of facts and named entity chain’s encodings. The following formulations show the transformations in the decoder component where the ground truth output sequence $y$ is passed to the sequence of transformer decoder layers and is used to initialize the first hidden layer $h^0$ of the decoder network. Note that we have a Masked Multi-head Self-Attention in the decoder network denoted by $MMHAtt$.

\begin{center}
$\tilde{h}^{l-2}_y := LayerNorm(h^{l-3}_y + MMHAtt(h^{l-3}_y))$ \newline
$\tilde{h}^{l-1}_y := LayerNorm(\tilde{h}^{l-2}_y + CrossAtt(\tilde{h}^{l-2}_y, {F}_K, \mathcal{E})$ \newline
$\tilde{h}^l_y := LayerNorm(\tilde{h}^{l-1}_y + CrossAtt(\tilde{h}^{l-1}_y, x))$ \newline
$h_y^l := LayerNorm(\tilde{h}^l_y + FFN(\tilde{h}^l_y))$ \newline    
\end{center}

\vspace{-3.5mm}
\section{Model Training}
All models are trained with a cross-entropy loss using backpropagation:

\begin{equation}
\mathcal L_{\theta} = -\frac{1}{n}\sum_{k=1}^n\mathcal P(t_k|t_{<k}, X, \mathcal E, F_K; \theta)    
\end{equation}

Where $X$ - the input sequence to be summarized; $\mathcal{E}$ - the named entities chain in the input sequence $X$; and ${F}_K$ - top-k facts extracted from biomedical KB; $\theta$ - model parameters.

We train each model using cross entropy loss to generate the ground truth summaries for the PubMed-50k dataset. We use the following hyperparameters setting: number of epochs to 5, fixed learning rate of 5e-5 with adam optimizer \cite{kingma2014adam},  batch size to 8, beam size of 5 with a length penalty \cite{wu2016google} $\alpha$ between the range of 0.6 and 1 \cite{liu2019text} at inference time; to deal with long-document summarization using the traiditional transformer encoder-decoder models, we split the source article into chunks of a maximum of 512 tokens and independently encode each chunk, after which we concatenate and project back to 768 dimension using a linear layer. The approach of splitting the long input sequence into smaller chunks of 512 tokens and then embedding independently is motivated by the recent work by \cite{deyoung2021ms2}. For Longformer-Encoder-Decoder (LED) \cite{beltagy2020longformer} and BigBird \cite{zaheer2020big}, however, we set the maximum length of the input sequence to 8192 tokens since they can deal with long input sequences without having to truncate; maximum output sequence length is set to 210 tokens following the experiments by \cite{cohan2018discourse}. To mitigate redundancy in the generated summaries, we enable trigram blocking \cite{paulus2017deep} during inference. For each backbone model, we use its base variant with 12 encoder and 12 decoder layers. The train/validation/test sizes for PubMed-50k is 50,000/5,000/5,000 and each model is trained using early stopping. A checkpoint of the model that performs the best (in terms of validation loss) on the validation set across different epochs is saved to the file system. All models are built and trained using PyTorch on NVIDIA Tesla T4 GPU. We perform model training experiments with different input guidance settings: input document only, input document + named entities chain, input document + named entities chain + knowledge facts. For our base summarization model, we experiment with five transformer-based encoder-decoder models and show that our entity-driven knowledge-aware approach enables us to achieve the best performance in entity-level factual consistency, N-gram novelty, and semantic equivalence while performing comparably on the commonly used ROUGE metrics.  At inference time, we experiment with two settings (w/o named entities, and w/ named entities).

\section{Experiments and Results}
While all models are trained using the PubMed-50k corpus, they are evaluated using a hold-out test set from the original PubMed dataset as well as the ICD-11-Summ-1000 corpus we curate. The experimental results are shown in Table-II through Table-IX. Results of evaluation w.r.t source articles reported are average results for both the ICD-11-Summ-1000 and PubMed corpora since the ICD-11 pseudo-extractive documents do not have a ground truth summary. For lexical (ROUGE) evaluation, we report ROUGE F1 scores \cite{lin2003automatic}. Similarly, here for evaluation conducted w.r.t source articles, the results reported are average results across the PubMed and ICD-11-Summ-1000 corpora. Entity-level factual accuracy \cite{nan2021entity} is measured in terms of precision, and recall w.r.t ground truth summary (for PubMed), and w.r.t source articles (for both PubMed and ICD-11-Summ-1000). Entity-level precision and recall w.r.t ground truth summaries are denoted with precision-target and recall-target; similarly, entity-level precision, and recall w.r.t the source article are designated with precision-source, and recall-source, respectively. The F1 score is the harmonic mean of the precision and recall for either case. For measuring semantic equivalence between generated summaries and ground truth summaries, we leverage BERTScore as proposed by \cite{zhang2019bertscore}; specifically, we use BioBERT for representing each token in a generated summary and the ground truth summary after which we perform pairwise cosine similarity as proposed in \cite{zhang2019bertscore}.  All experimental results are reported in percentages. The average full text length of input source articles in PubMed-50k is 3,224 words and the average abstract length is 218 words, while for ICD-11-Summ-1000, the average length of an extractive pseudo-doc (i.e., input source article) is 4816 words.

\begin{table}
\captionsetup{font=scriptsize}
\begin{center}

\arrayrulecolor{black}

\begin{adjustbox}{width=0.5\textwidth}
\begin{tabular}{|c|c|c|c|c|}


\arrayrulecolor{black}\cline{1-1}\arrayrulecolor{black}\cline{2-5}
\textbf{Backbone Model} & \textbf{Training Config ($K$=3)} & \textbf{R-1} & \textbf{R-2} & \textbf{R-L} \\ 
\arrayrulecolor{black}\cline{1-1}\arrayrulecolor{black}\cline{2-5}
\multirow{3}{*}{T5} & T5 Vanilla (Baseline) & \textbf{31.333} & \textbf{12.821} & \textbf{29.018} \\ 
\cline{2-5}
 & T5 w/ named entities (Ours) & 29.915 & 11.352 & 27.667 \\ 
\cline{2-5}
 & T5 w/ named entities /w facts - EFAS (Ours) & 28.643 & 11.286 & 26.591 \\ 
\hline
\multirow{3}{*}{BART} & BART Vanilla (Baseline) & \textbf{34.214} & \textbf{13.830} & \textbf{31.545} \\ 
\cline{2-5}
 & BART w/ named entities (Ours) & 32.377 & 11.733 & 29.910 \\ 
\cline{2-5}
 & BART w/ named entities /w facts - EFAS (Ours) & 31.283 & 10.528 & 28.174 \\ 
\hline
\multirow{3}{*}{Pegasus} & Pegasus Vanilla (Baseline) & 28.851 & 11.274 & 26.859 \\ 
\cline{2-5}
 & Pegasus w/ named entities (Ours) & 30.365 & 11.483 & 28.003 \\ 
\cline{2-5}
 & Pegasus w/ named entities /w facts - EFAS (Ours) & \textbf{30.872} & \textbf{12.031} & \textbf{28.263} \\
\hline

\multirow{3}{*}{BigBird} & BigBird Vanilla (Baseline) & \textbf{35.426} & \textbf{13.801} & \textbf{32.537} \\ 
\cline{2-5}
 & BigBird w/ named entities (Ours) & 33.491 & 12.362 & 30.184 \\ 
\cline{2-5}
 & BigBird w/ named entities /w facts - EFAS (Ours) & 31.936 & 13.162 & 28.730 \\
\hline

\multirow{3}{*}{LED} & LED Vanilla (Baseline) & \textbf{36.218} & \textbf{14.173} & \textbf{32.862} \\ 
\cline{2-5}
 & LED w/ named entities (Ours) & 33.734 & 13.825 & 30.614 \\ 
\cline{2-5}
 & LED w/ named entities /w facts - EFAS (Ours) & 33.283 & 13.582 & 29.038 \\
\hline

\end{tabular}
\end{adjustbox}
\arrayrulecolor{black}
\end{center}
\caption{Lexical (ROUGE) Evaluation w.r.t Ground Truth Summary (\emph{vanilla input @ inference time}). The input in this experimental setting is the \emph{raw input article to be summarized (i.e., w/o named entity chain)}}

\end{table}

\begin{table}
\captionsetup{font=scriptsize}

\begin{center}
\centering

\arrayrulecolor{black}
\begin{adjustbox}{width=0.5\textwidth}
\begin{tabular}{!{\color{black}\vrule}c!{\color{black}\vrule}c!{\color{black}\vrule}c!{\color{black}\vrule}c!{\color{black}\vrule}c!{\color{black}\vrule}} 
\arrayrulecolor{black}\cline{1-1}\arrayrulecolor{black}\cline{2-5}
\multirow{2}{*}{\textbf{Backbone Model}} & \multirow{2}{*}{\textbf{Training Config ($K$=3)}} & \multicolumn{3}{c!{\color{black}\vrule}}{\textbf{Entity-level Factual Consistency}} \\ 
\cline{3-5}
 &  & \textbf{Precision-target} & \textbf{Recall-target} & \textbf{F1 score-target} \\ 
\arrayrulecolor{black}\cline{1-1}\arrayrulecolor{black}\cline{2-5}
\multirow{3}{*}{{T5}} & T5 Vanilla (Baseline) & 27.008 & \textbf{21.175} & \textbf{23.738} \\ 
\cline{2-5}
 & T5 w/ named entities (Ours) & \textbf{27.564} & 19.246 & 22.666 \\ 
\cline{2-5}
 & T5 w/ named entities /w facts - EFAS (Ours) & 27.329 & 19.136 & 22.510 \\ 
\hline
\multirow{3}{*}{BART} & BART Vanilla (Baseline) & \textbf{28.315} & \textbf{20.404} & \textbf{23.718} \\ 
\cline{2-5}
 & BART w/ named entities (Ours) & 27.949 & 19.105 & 22.695 \\ 
\cline{2-5}
 & BART w/ named entities /w facts - EFAS (Ours) & 27.241 & 18.792 & 22.241 \\ 
\hline
\multirow{3}{*}{Pegasus} & Pegasus Vanilla (Baseline) & 17.911 & 20.212 & 18.992 \\ 
\cline{2-5}
 & Pegasus w/ named entities (Ours) & 22.950 & 21.335 & 22.113 \\ 
\cline{2-5}
 & Pegasus w/ named entities /w facts - EFAS (Ours) & \textbf{23.572} & \textbf{22.956} & \textbf{23.260} \\
\hline

\multirow{3}{*}{BigBird} & BigBird Vanilla (Baseline) & 16.523 & 19.384 & 17.840 \\ 
\cline{2-5}
 & BigBird w/ named entities (Ours) & 23.273 & 21.831 & 22.529 \\ 
\cline{2-5}
 & BigBird w/ named entities /w facts - EFAS (Ours) & \textbf{25.317} & \textbf{23.839} & \textbf{24.556} \\
\hline

\multirow{3}{*}{LED} & LED Vanilla (Baseline) & 17.830 & 20.173 & 18.929 \\ 
\cline{2-5}
 & LED w/ named entities (Ours) & 24.528 & 22.573 & 23.510 \\ 
\cline{2-5}
 & LED w/ named entities /w facts - EFAS (Ours) & \textbf{26.827} & \textbf{25.322} & \textbf{26.053} \\
\hline

\end{tabular}
\end{adjustbox}
\arrayrulecolor{black}
\end{center}
\caption{Entity-level Factual Consistency Evaluation w.r.t Ground Truth Summary (\emph{vanilla input @ inference time}). The input in this experimental setting is the \emph{raw input article to be summarized (i.e., w/o named entity chain)}}
\end{table}

\begin{table}
\captionsetup{font=scriptsize}
\begin{center}
\centering
\arrayrulecolor{black}
\begin{adjustbox}{width=0.5\textwidth}
\begin{tabular}{!{\color{black}\vrule}c!{\color{black}\vrule}c!{\color{black}\vrule}c!{\color{black}\vrule}c!{\color{black}\vrule}c!{\color{black}\vrule}} 
\arrayrulecolor{black}\cline{1-1}\arrayrulecolor{black}\cline{2-5}
\multirow{2}{*}{\textbf{Backbone Model}} & \multirow{2}{*}{\textbf{Training Config ($K$=3)}} & \multicolumn{3}{c!{\color{black}\vrule}}{\textbf{Entity-level Factual Consistency}} \\ 
\cline{3-5}
 &  & \textbf{Precision-source} & \textbf{Recall-source} & \textbf{F1 score-source} \\ 
\arrayrulecolor{black}\cline{1-1}\arrayrulecolor{black}\cline{2-5}
\multirow{3}{*}{T5} & T5 Vanilla (Baseline) & \textbf{55.076} & \textbf{7.976} & \textbf{13.934} \\ 
\cline{2-5}
 & T5 w/ named entities (Ours) & 54.015 & 7.232 & 12.756 \\ 
\cline{2-5}
 & T5 w/ named entities /w facts - EFAS (Ours) & 53.284 & 6.275 & 11.228 \\ 
\hline
\multirow{3}{*}{BART} & BART Vanilla (Baseline) & 58.592 & \textbf{5.623} & \textbf{10.261} \\ 
\cline{2-5}
 & BART w/ named entities (Ours) & 60.422 & 5.361 & 9.848 \\ 
\cline{2-5}
 & BART w/ named entities /w facts - EFAS (Ours) & \textbf{61.593} & 4.739 & 8.801 \\ 
\hline
\multirow{3}{*}{Pegasus} & Pegasus Vanilla (Baseline) & 33.821 & 7.401 & 12.144 \\ 
\cline{2-5}
 & Pegasus w/ named entities (Ours) & 46.757 & 7.743 & 13.286 \\ 
\cline{2-5}
 & Pegasus w/ named entities /w facts - EFAS (Ours) & \textbf{48.387} & \textbf{8.263} & \textbf{14.116} \\
\hline

\multirow{3}{*}{BigBird} & BigBird Vanilla (Baseline) & 34.288 & 9.261 & 14.583 \\ 
\cline{2-5}
 & BigBird w/ named entities (Ours) & 48.283 & 8.625 & 14.636 \\ 
\cline{2-5}
 & BigBird w/ named entities /w facts - EFAS (Ours) & \textbf{48.572} & \textbf{9.583} & \textbf{16.008} \\
\hline

\multirow{3}{*}{LED} & LED Vanilla (Baseline) & 59.361 & 6.731 & 12.091 \\ 
\cline{2-5}
 & LED w/ named entities (Ours) & 62.479 & 6.382 & 11.581 \\ 
\cline{2-5}
 & LED w/ named entities /w facts - EFAS (Ours) & \textbf{63.731} & \textbf{6.821} & \textbf{12.323} \\
\hline

\end{tabular}
\end{adjustbox}
\arrayrulecolor{black}
\end{center}
\caption{Entity-level Factual Consistency \emph{w.r.t source article}. The input in this experimental setting is the raw input article to be summarized @ inference time (i.e., w/o named entity chain).}
\end{table}


\begin{table}
\captionsetup{font=scriptsize}
\begin{center}
\centering
\arrayrulecolor{black}
\begin{adjustbox}{width=0.5\textwidth}
\begin{tabular}{|c|c|c|c|c|}

\hline

\textbf{Backbone Model} & \textbf{Training Config ($k$=3)} & \textbf{R-1} & \textbf{R-2} & \textbf{R-L} \\ 
\hline
\multirow{3}{*}{T5} & T5 Vanilla (Baseline) & 29.837 & 11.386 & 27.493 \\ 
\cline{2-5}
 & T5 w/ named entities (Ours) & \textbf{32.183} & \textbf{13.725} & \textbf{29.398} \\ 
\cline{2-5}
 & T5 w/ named entities /w facts - EFAS (Ours) & 29.372 & 9.682 & 28.275 \\ 
\hline
\multirow{3}{*}{BART} & BART Vanilla (Baseline) & 34.762 & 12.592 & 29.387 \\ 
\cline{2-5}
 & BART w/ named entities (Ours) & \textbf{35.281} & \textbf{12.938} & \textbf{31.276} \\ 
\cline{2-5}
 & BART w/ named entities /w facts - EFAS (Ours) & 33.731 & 11.923 & 30.285 \\ 
\hline
\multirow{3}{*}{Pegasus} & Pegasus Vanilla (Baseline) & 26.592 & 10.052 & 24.386 \\ 
\cline{2-5}
 & Pegasus w/ named entities (Ours) & 32.562 & \textbf{13.864} & 30.174 \\ 
\cline{2-5}
 & Pegasus w/ named entities /w facts - EFAS (Ours) & \textbf{33.824} & 13.841 & \textbf{30.639} \\
\hline

\multirow{3}{*}{BigBird} & BigBird Vanilla (Baseline) & 28.174 & 11.371 & 25.692 \\ 
\cline{2-5}
 & BigBird w/ named entities (Ours) & 32.281 & \textbf{14.263} & \textbf{31.863} \\ 
\cline{2-5}
 & BigBird w/ named entities /w facts - EFAS (Ours) & \textbf{34.728} & 13.264 & 31.752 \\
\hline

\multirow{3}{*}{LED} & LED Vanilla (Baseline) & 34.265 & 10.826 & 26.173 \\ 
\cline{2-5}
 & LED w/ named entities (Ours) & \textbf{36.840} & 13.773 & \textbf{32.156} \\ 
\cline{2-5}
 & LED w/ named entities /w facts - EFAS (Ours) & 34.927 & \textbf{14.003} & 30.851 \\
\hline

\end{tabular}
\end{adjustbox}
\arrayrulecolor{black}
\end{center}
\caption{Lexical (ROUGE) Evaluation w.r.t Ground Truth Summary (\emph{input article + named entity chain @ inference time}); i.e., the input in this experimental setting is the \emph{raw input article to be summarized with the named entities (i.e., w/ named entity chain)}.}
\end{table}


\begin{table}
\captionsetup{font=scriptsize}
\begin{center}
\centering
\arrayrulecolor{black}
\begin{adjustbox}{width=0.5\textwidth}
\begin{tabular}{|c|c|c|c|c|}

\hline
\multirow{2}{*}{\textbf{Backbone Model}} & \multirow{2}{*}{\textbf{Training Config ($K$=3)}} & 

\multicolumn{3}{|c|}{\textbf{Entity-level Factual Consistency}}\\

\cline{3-5}
 &  & \textbf{Precision-target} & \textbf{Recall-target} & \textbf{F1 score-target} \\ 
\hline
\multirow{3}{*}{T5} & T5 Vanilla (Baseline) & 26.194 & 19.759 & 22.526 \\ 
\cline{2-5}
 & T5 w/ named entities (Ours) & \textbf{29.826} & \textbf{22.952} & \textbf{25.941} \\ 
\cline{2-5}
 & T5 w/ named entities /w facts - EFAS (Ours) & 28.582 & 20.738 & 24.036 \\ 
\hline
\multirow{3}{*}{BART} & BART Vanilla (Baseline) & 26.581 & 18.381 & 21.733 \\ 
\cline{2-5}
 & BART w/ named entities (Ours) & \textbf{27.949} & \textbf{19.105} & \textbf{22.696} \\ 
\cline{2-5}
 & BART w/ named entities /w facts - EFAS (Ours) & 27.241 & 18.792 & 22.241 \\ 
\hline
\multirow{3}{*}{Pegasus} & Pegasus Vanilla (Baseline) & 15.386 & 19.382 & 17.154 \\ 
\cline{2-5}
 & Pegasus w/ named entities (Ours) & 24.638 & 23.529 & 24.071 \\ 
\cline{2-5}
 & Pegasus w/ named entities /w facts - EFAS (Ours) & \textbf{25.498} & \textbf{24.374} & \textbf{24.923} \\
\hline

\multirow{3}{*}{BigBird} & BigBird Vanilla (Baseline) & 15.942 & 19.873 & 17.692 \\ 
\cline{2-5}
 & BigBird w/ named entities (Ours) & 26.315 & \textbf{24.728} & \textbf{25.497} \\ 
\cline{2-5}
 & BigBird w/ named entities /w facts - EFAS (Ours) & \textbf{26.638} & 24.163 & 25.340 \\
\hline

\multirow{3}{*}{LED} & LED Vanilla (Baseline) & 17.284 & 20.692 & 18.835 \\ 
\cline{2-5}
 & LED w/ named entities (Ours) & \textbf{28.173} & 25.866 & \textbf{26.970} \\ 
\cline{2-5}
 & LED w/ named entities /w facts - EFAS (Ours) & 26.116 & \textbf{26.830} & 26.468 \\
\hline

\end{tabular}
\end{adjustbox}
\arrayrulecolor{black}
\end{center}
\caption{Entity-level Factual Consistency w.r.t Ground Truth Summary. The input in this experimental setting is the \emph{raw input article to be summarized with the named entities (i.e., w/ named entity chain)} @ inference time.}
\end{table}

\begin{table}
\captionsetup{font=scriptsize}
\centering
\begin{center}
\arrayrulecolor{black}
\begin{adjustbox}{width=0.5\textwidth}
\begin{tabular}{|c|c|c|c|c|}

\hline
\multirow{2}{*}{\textbf{Backbone Model}} & \multirow{2}{*}{\textbf{Training Config ($K$=3)}} & 

\multicolumn{3}{|c|}{\textbf{Entity-level Factual Consistency}}\\

\cline{3-5}
 &  & \textbf{Precision-source} & \textbf{Recall-source} & \textbf{F1 score-source} \\ 
\hline
\multirow{3}{*}{T5} & T5 Vanilla (Baseline) & 52.183 & 5.792 & 10.427 \\ 
\cline{2-5}
 & T5 w/ named entities (Ours) & \textbf{56.803} & \textbf{10.816} & \textbf{18.172} \\ 
\cline{2-5}
 & T5 w/ named entities /w facts - EFAS (Ours) & 55.728 & 8.629 & 14.944 \\ 
\hline
\multirow{3}{*}{BART} & BART Vanilla (Baseline) & 56.611 & 5.031 & 9.241 \\ 
\cline{2-5}
 & BART w/ named entities (Ours) & \textbf{62.385} & \textbf{7.284} & \textbf{13.045} \\ 
\cline{2-5}
 & BART w/ named entities /w facts - EFAS (Ours) & 61.938 & 6.382 & 11.572 \\ 
\hline
\multirow{3}{*}{Pegasus} & Pegasus Vanilla (Baseline) & 31.492 & 6.792 & 11.174 \\ 
\cline{2-5}
 & Pegasus w/ named entities (Ours) & 48.389 & 8.396 & 14.309 \\ 
\cline{2-5}
 & Pegasus w/ named entities /w facts - EFAS (Ours) & \textbf{48.964} & \textbf{9.491} & \textbf{15.900} \\
\hline

\multirow{3}{*}{BigBird} & BigBird Vanilla (Baseline) & 31.882 & 8.177 & 13.016 \\ 
\cline{2-5}
 & BigBird w/ named entities (Ours) & 48.733 & 9.267 & 15.573 \\ 
\cline{2-5}
 & BigBird w/ named entities /w facts - EFAS (Ours) & \textbf{50.373} & \textbf{11.274} & \textbf{18.424} \\
\hline

\multirow{3}{*}{LED} & LED Vanilla (Baseline) & 58.316 & 6.472 & 11.651 \\ 
\cline{2-5}
 & LED w/ named entities (Ours) & 63.722 & \textbf{8.537} & \textbf{15.057} \\ 
\cline{2-5}
 & LED w/ named entities /w facts - EFAS (Ours) & \textbf{65.180} & 8.374 & 14.841 \\ 
\hline

\end{tabular}
\end{adjustbox}
\arrayrulecolor{black}
\end{center}
\caption{Entity-level Factual Consistency \emph{w.r.t input source article} (\emph{input article + named entity chain @ inference time}); i.e., the input in this experimental setting is the \emph{raw input article to be summarized with the named entities (i.e., w/ named entity chain)}.}
\end{table}



\begin{table}
\captionsetup{font=scriptsize}
\centering
\arrayrulecolor{black}
\begin{adjustbox}{width=0.5\textwidth}
\begin{tabular}{!{\color{black}\vrule}c!{\color{black}\vrule}c!{\color{black}\vrule}c!{\color{black}\vrule}c!{\color{black}\vrule}} 
\arrayrulecolor{black}\cline{1-1}\arrayrulecolor{black}\cline{2-4}
\multirow{2}{*}{\textbf{Backbone Model}} & \multirow{2}{*}{\textbf{Training Configuration ($K$=3)}} & \multicolumn{2}{c!{\color{black}\vrule}}{\textbf{N-gram Novelty}} \\ 
\cline{3-4}
 &  & \textbf{w/o named entities} & \textbf{w/ named entities} \\ 
\arrayrulecolor{black}\cline{1-1}\arrayrulecolor{black}\cline{2-4}
\multirow{3}{*}{T5} & T5 Vanilla (Baseline) & 52.930 & 49.699 \\ 
\cline{2-4}
 & T5 w/ named entities (Ours) & 50.079 & 50.967 \\ 
\cline{2-4}
 & T5 w/ named entities /w facts - EFAS (Ours) & \textbf{53.817} & \textbf{52.841} \\ 
\hline
\multirow{3}{*}{BART} & BART Vanilla (Baseline) & 54.816 & 54.997 \\ 
\cline{2-4}
 & BART w/ named entities (Ours) & 54.959 & 57.811 \\ 
\cline{2-4}
 & BART w/ named entities /w facts - EFAS (Ours) & \textbf{57.360} & \textbf{61.370} \\ 
\hline
\multirow{3}{*}{Pegasus} & Pegasus Vanilla (Baseline) & 51.260 & 50.035 \\ 
\cline{2-4}
 & Pegasus w/ named entities (Ours) & 52.558 & 51.269 \\ 
\cline{2-4}
 & Pegasus w/ named entities /w facts - EFAS (Ours) & \textbf{54.621} & \textbf{52.702} \\
\hline

\multirow{3}{*}{BigBird} & BigBird Vanilla (Baseline) & 49.783 & 51.374 \\ 
\cline{2-4}
 & BigBird w/ named entities (Ours) & 52.729 & \textbf{54.836} \\ 
\cline{2-4}
 & BigBird w/ named entities /w facts - EFAS (Ours) & \textbf{53.661} & 53.827 \\
\hline

\multirow{3}{*}{LED} & LED Vanilla (Baseline) & 53.732 & 53.288 \\ 
\cline{2-4}
 & LED w/ named entities (Ours) & 55.826 & 58.637 \\ 
\cline{2-4}
 & LED w/ named entities /w facts - EFAS (Ours) & \textbf{59.283} & \textbf{61.482} \\ 
\hline

\end{tabular}
\end{adjustbox}
\arrayrulecolor{black}
\caption{N-gram Novelty w.r.t source articles w/o and w/ named entity chain during inference.}
\end{table}


\begin{table}
\captionsetup{font=scriptsize}
\centering
\arrayrulecolor{black}
\begin{adjustbox}{width=0.5\textwidth}
\begin{tabular}{!{\color{black}\vrule}c!{\color{black}\vrule}c!{\color{black}\vrule}c!{\color{black}\vrule}c!{\color{black}\vrule}} 
\arrayrulecolor{black}\cline{1-1}\arrayrulecolor{black}\cline{2-4}
\multirow{2}{*}{\textbf{Backbone Model}} & \multirow{2}{*}{\textbf{Training Configuration ($K$=3)}} & \multicolumn{2}{c!{\color{black}\vrule}}{\textbf{BioBERTScore}} \\ 
\cline{3-4}
 &  & \textbf{w/o named entities} & \textbf{w/ named entities} \\ 
\arrayrulecolor{black}\cline{1-1}\arrayrulecolor{black}\cline{2-4}
\multirow{3}{*}{T5} & T5 Vanilla (Baseline) & 52.269 & 51.682 \\ 
\cline{2-4}
 & T5 w/ named entities (Ours) & 51.868 & 52.739 \\ 
\cline{2-4}
 & T5 w/ named entities /w facts - EFAS (Ours) & \textbf{53.162} & \textbf{54.164} \\ 
\hline
\multirow{3}{*}{BART} & BART Vanilla (Baseline) & 51.799 & 50.283 \\ 
\cline{2-4}
 & BART w/ named entities (Ours) & 51.783 & \textbf{53.618} \\ 
\cline{2-4}
 & BART w/ named entities /w facts - EFAS (Ours) & \textbf{52.072} & 51.472 \\ 
\hline
\multirow{3}{*}{Pegasus} & Pegasus Vanilla (Baseline) & 53.168 & 51.381 \\ 
\cline{2-4}
 & Pegasus w/ named entities (Ours) & 53.401 & \textbf{55.761} \\ 
\cline{2-4}
 & Pegasus w/ named entities /w facts - EFAS (Ours) & \textbf{54.382} & 55.263 \\
\hline

\multirow{3}{*}{BigBird} & BigBird Vanilla (Baseline) & 55.271 & 53.620 \\ 
\cline{2-4}
 & BigBird w/ named entities (Ours) & \textbf{56.813} & 54.271 \\ 
\cline{2-4}
 & BigBird w/ named entities /w facts - EFAS (Ours) & 56.372 & 55.088 \\
\hline

\multirow{3}{*}{LED} & LED Vanilla (Baseline) & 53.732 & 52.427 \\ 
\cline{2-4}
 & LED w/ named entities (Ours) & \textbf{54.163} & 55.791 \\ 
\cline{2-4}
 & LED w/ named entities /w facts - EFAS (Ours) & 53.814 & \textbf{57.284} \\ 
\hline

\end{tabular}
\end{adjustbox}
\arrayrulecolor{black}
\caption{Semantic Equivalence (BioBERTScore \cite{zhang2019bertscore}) w.r.t ground truth summaries w/o and w/ named entity chain during inference. Since we are using BioBERT for representation learning, we refer to the metric as BioBERTScore, a variant of BERTScore.}
\end{table}


\section{Ablation Studies}
To assess the impact of facts mined on the quality of summaries generated, we conduct an ablation study where we experiment with different values of $K$ in top-k for the backbone models. Figure-4 shows results of ablation to assess precision-source and recall-target. Since we want to minimize entity hallucination which is measured in terms of precision-source and want to maximize the number of entities in the ground truth summary that are retrieved in the generated summary as measured by recall-target, we report the impact of different values of $K$ for these two metrics. As shown in the two plots, precision-source and recall-target consistently improve as we retrieve more relevant facts from the biomedical knowledge bases and train our models. 

\vspace{-1.2mm}

\begin{figure}
\centering
\captionsetup{font=scriptsize}
  \begin{subfigure}[b]{0.24\textwidth}
    \includegraphics[width=\textwidth]{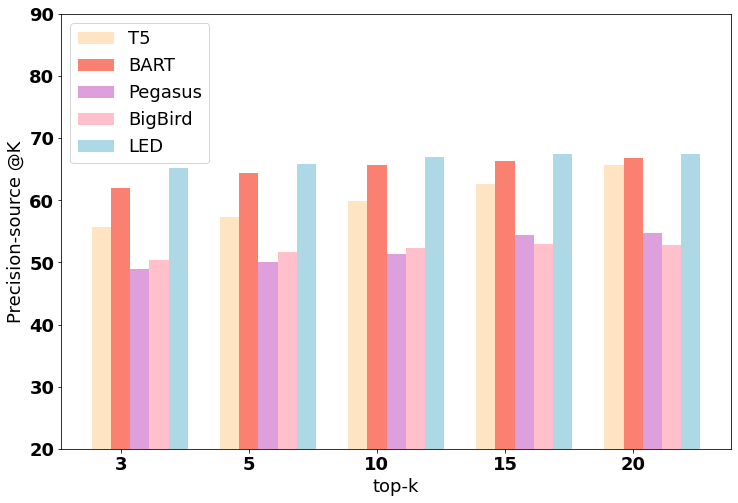}
    \label{fig:1}
  \end{subfigure}
 \hfill
  \begin{subfigure}[b]{0.24\textwidth}
    \includegraphics[width=\textwidth]{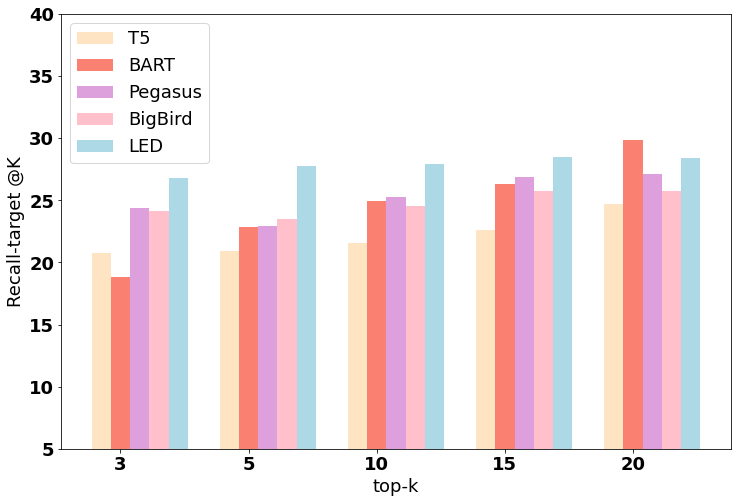}
    \label{fig:2}
  \end{subfigure}

\vspace{-4mm}
\caption{Precision-source and Recall-target for different values of $K$. As can be seen, both metrics slightly increase as we increase the number of facts used to train a model. Note that this evaluation is done with source article and named entity chain passed to the trained models at inference time.}
\end{figure}

\section{Discussion of Results}


From the results reported in the previous section, we generally see entity-level factual consistency (particularly, precision-source, and recall-target) improve when a model is trained with named entities and/or facts included as an additional signal in the training with the same objective of generating the ground truth summary using cross-entropy loss. The addition of more facts further improves entity-level factual consistency as shown in Figure-4. Further, we notice N-gram novelty improves with our proposed framework for the five backbone models. Semantic equivalence generally improves when named entities and/or facts are included during training for all models.  Thus, the corresponding entries for the various models and training configurations show improvement in semantic based scores. The ROUGE scores, however, drop slightly from when there is no additional context at training or inference time. The drop in ROUGE is a result of augmenting the models with facts from background knowledge bases which in turn leads to higher N-gram novelty. Thus, the proposed framework enables us to achieve better abstractive scores in terms of entity-level factual consistency, paraphrasing and semantic equivalence. 

\section{Conclusions and Future Work}
In this study, we proposed a framework to integrate named entities in a source article and facts extracted from biomedical knowledge bases pertaining to the named entities in transformer-based encoder-decoder models and applied to the task of abstractive summarization of biomedical literature. Through extensive experiments, we showed the proposed approach improves the semantics of generated summaries in terms of entity-level factual consistency and semantic equivalence while generating novel words. For future steps, we plan to jointly train the knowledge-retriever and the knowledge-guided abstractive summarizer in an end-to-end fashion. While our current architecture optimizes a single cost function given different input signals, we plan to augment the existing framework using multi-objective optimization to further enhance the factual accuracy and semantic equivalence of generated summaries using different cost functions during training.

\bibliographystyle{IEEEtran} 
\bibliography{bibliography.bib}

\end{document}